\documentclass[conference]{IEEEtran}
\IEEEoverridecommandlockouts
\usepackage{cite}
\usepackage{amsmath,amssymb,amsfonts}
\usepackage{algorithmic}
\usepackage{graphicx}
\usepackage{textcomp}
\usepackage{xcolor}
\usepackage{balance}
\usepackage{url}
\usepackage{siunitx}
\usepackage{hyperref}
\usepackage{subcaption}
\def\BibTeX{{\rm B\kern-.05em{\sc i\kern-.025em b}\kern-.08em
    T\kern-.1667em\lower.7ex\hbox{E}\kern-.125emX}}
\begin{document}

\title{Unsupervised Transcript-assisted Video Summarization and Highlight Detection\\
}

\author{
\IEEEauthorblockN{Spyros Barbakos$^1$, Charalampos Antoniadis$^2$, Gerasimos Potamianos$^1$, Gianluca Setti$^2$}
\IEEEauthorblockA{\small$^{1}$University of Thessaly, Volos, Greece\\
$^{2}$King Abdullah University of Science and Technology (KAUST), Thuwal, Saudi Arabia
}}

\maketitle

\begin{abstract}
Video consumption is a key part of daily life, but watching entire videos can be tedious. To address this, researchers have explored video summarization and highlight detection to identify key video segments. While some works combine video frames and transcripts, and others tackle video summarization and highlight detection using Reinforcement Learning (RL), no existing work, to the best of our knowledge, integrates both modalities within an RL framework. In this paper, we propose a multimodal pipeline that leverages video frames and their corresponding transcripts to generate a more condensed version of the video and detect highlights using a modality fusion mechanism. The pipeline is trained within an RL framework, which rewards the model for generating diverse and representative summaries while ensuring the inclusion of video segments with meaningful transcript content. The unsupervised nature of the training allows for learning from large-scale unannotated datasets, overcoming the challenge posed by the limited size of existing annotated datasets. 
Our experiments show that using the transcript in video summarization and highlight detection achieves superior results compared to relying solely on the visual content of the video.

\end{abstract}


\section{Introduction}

As digital media continues to dominate daily life, videos have become a primary source of information and entertainment. However, watching full-length videos can be very time-consuming. To address this challenge, researchers have focused on video summarization and highlight detection to automatically identify the most informative and engaging segments of videos. Specifically, video summarization generates a shorter, coherent version of the original video, while highlight detection selects the most important frames without necessarily preserving overall coherence.

Recent studies~\cite{he2023a2summ, clover, susinet, UMT, clipit} have proposed deep-learning models trained on manually annotated datasets in a supervised manner to generate video summaries. However, the key limitation of this approach is the scarcity of annotated video summarization datasets, as human annotation is both costly and labor-intensive  \cite{summe}, \cite{TVSUM}.

To overcome this limitation, researchers have turned to models that use Reinforcement Learning (RL) or weak supervision to identify important video segments and generate summaries. In~\cite{vsummReinforce}, researchers introduced formulas to quantify the diversity and representativeness of a video summary, considering their score as a reward signal to train an LSTM-based summarization network via RL. Moreover, in \cite{casum}, a self-attention mechanism is applied to non-overlapping blocks to estimate frame importance, assessing their uniqueness and diversity. In \cite{tldw}, researchers introduce a weakly supervised framework that exploits cross-video commonalities to generate synthetic annotations that are used to train a Transformer-based summarizer.

Another key aspect of video summarization research is how models process video inputs. A video consists of three fundamental components: visual frames, speech (transcripts), and audio. Some models rely solely on visual frames (unimodal ones)~\cite{susinet},~\cite{vsummReinforce}, while others incorporate both frames and transcripts (multimodal ones) or enhance datasets by generating descriptive captions for each frame~\cite{he2023a2summ},~\cite{clover},~\cite{UMT},~\cite{clipit}.

Extensive research has been conducted on unimodal and multimodal models trained using supervised learning, as well as on unimodal models trained in an unsupervised manner. However, unsupervised multimodal models remain largely unexplored. Exploring this under-researched area could enable models to leverage multimodal information from vast amounts of raw video data, leading to more robust summarization systems.

In this paper, we propose a deep learning solution within an RL framework that leverages both the visual content of a video and its transcript, i.e., the information conveyed through spoken language. Our approach employs an advanced Transformer model for frame-text fusion, ensuring that frames attend only to their corresponding sentences. Additionally, we integrate Large Language Models (LLMs) to evaluate the significance of the video transcript and use this score as an additional reward alongside the visual reward.

\section{Proposed Approach}
\label{proposed_approach}

\subsection{Overview}
\label{overview}
We formulate video summarization as an RL problem, where a deep learning model, acting as the policy $\pi$, determines which frames to include in the summary. Our RL approach defines a reward function that evaluates a video summary $\mathcal{S}$ not only based on its visual content \cite{vsummReinforce} but also on the significance of the transcript corresponding to the selected frames in the summary.

\begin{figure}
    \centering
    \includegraphics[width=\linewidth]{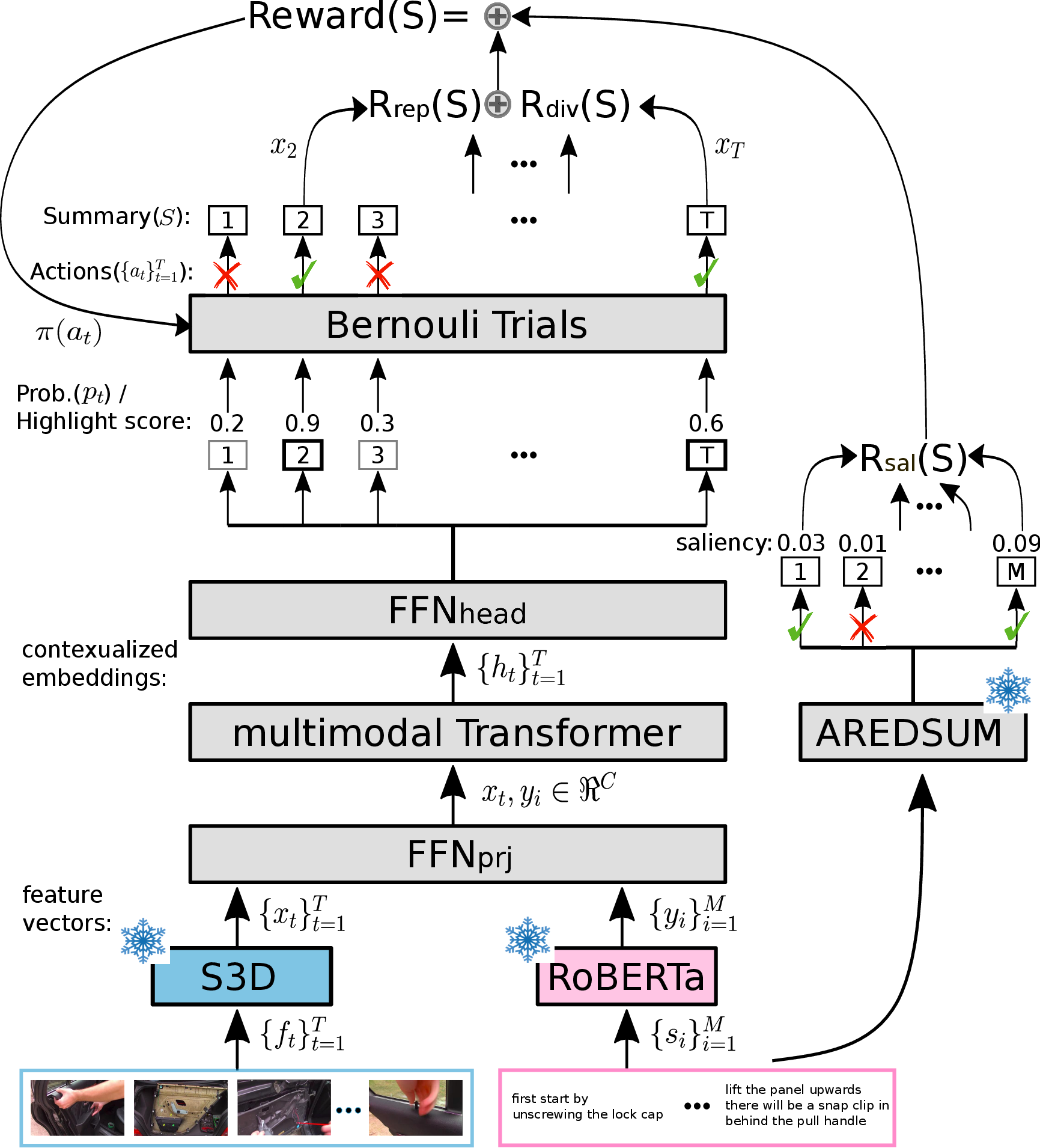}
    \caption{Block diagram of the proposed RL-based deep-learning solution to video summarization and highlight detection. The snowflakes indicate that the models are only used for inference.}
    \label{fig:overview}
\end{figure}

In Fig.~\ref{fig:overview}, we present an overview of our RL-based deep-learning solution. The proposed pipeline first extracts feature vectors $\{x_t\}_{t=1}^T$, where $T$ is the total number of frames and $\{y_i\}_{i=1}^M$, where $M$ is the total number of sentences in the transcript, from the video frames $\{f_t\}_{t=1}^T$ and the sentences $\{s_i\}_{i=1}^M$ in the transcript using the S3D \cite{s3d} and the RoBERTa \cite{roberta} pre-trained models, respectively. Then, a feedforward neural network ($\text{FFN}_\text{{prj}}$) projects these feature vectors into a common $C$-dimensional embedding space. After that, these embeddings are provided as input to a multimodal Transformer model that extracts contextualized embeddings ($\{h_t\}_{t=1}^T$) for each frame.
A final feedforward network ($\text{FFN}_\text{{head}}$) with a sigmoid activation assigns a probability $p_t$ to each frame, representing its likelihood of inclusion in the summary or its highlight score. 

The training of the model relies on the maximization of the reward that we define in the RL framework.
The visual reward is computed from the frame embeddings, while the transcript reward is derived using a pre-trained model~\cite{AREDSUM} that scores each sentence in the transcript based on its saliency.

During inference, for the case of highlight detection, we treat the $p_t$ probabilities as the highlight scores, while for the case of video summarization, we segment the video into shots based on visual coherence, and a greedy algorithm selects the shots for inclusion in the summary, based on the averaged $p_t$ values within the shot.

\subsection{Reward Functions}

The reward function $R(\mathcal{S})$ used in our RL framework is defined:
\begin{equation}
    R(\mathcal{S}) = R_{div}(\mathcal{S}) + R_{rep}(\mathcal{S}) + R_{sal}(\mathcal{S}),
\end{equation}

\noindent
where $R_{div}(\mathcal{S})$ and $R_{rep}(\mathcal{S})$ evaluate the diversity and representativeness of the selected frames in the summary, respectively, while $R_{sal}(\mathcal{S})$ measures the saliency of the transcript corresponding to these selected frames.

\subsubsection{Diversity Reward ($R_{div}$)}
The diversity reward relies on the assumption that a high-quality summary should contain a wide range of distinct frames from the initial video. We define the \textit{Diversity Reward} $R_{div}$ as the average pairwise dissimilarity between the selected frames~\cite{vsummReinforce}:
\begin{equation} \label{eq:diversityreward}
R_{div}(\mathcal{S}) = \frac{1}{|\mathcal{S}| (|\mathcal{S}| - 1)} \sum_{t \in \mathcal{S}} \sum_{\substack{t^\prime \in \mathcal{S} \\ t^\prime \neq t}} d (x_t, x_{t^\prime}),
\end{equation}
where $d(\cdot, \cdot)$ is the dissimilarity function given by:
\begin{equation}
d (x_t, x_{t^\prime}) = 1 - \frac{x_t^T x_{t^\prime}}{||x_t||_2 ||x_{t^\prime}||_2},
\end{equation}

\noindent
and $x_t$ are the frame S3D feature vectors \cite{s3d}. 

The diversity reward function promotes the inclusion of dissimilar frames in the summary, i.e., ensuring that $x_{t^\prime}$ and $x_t$ are as orthogonal as possible. Moreover, since two frames that are far apart in the sequence - let’s say by $\lambda$ frames - can be crucial to the overall storyline, we set $d(x_t, x_{t^\prime}) = 1$ by default whenever $|t-t^\prime|>\lambda$\cite{vsummReinforce}.

\subsubsection{Representativeness Reward ($R_{rep}$)}

Frames that look similar are expected to have corresponding feature vectors that are close to each other, forming clusters in the feature vector space. The \textit{Representativeness Reward} $R_{rep}$ is defined as follows~\cite{vsummReinforce}:

\begin{equation} \label{eq:represenativeness_reward}
R_{rep}(\mathcal{S}) = \exp  \Big(- \frac{1}{T} \sum_{t=1}^T \min_{t^\prime \in \mathcal{S}} || x_t - x_{t^\prime} ||_2 \Big).
\end{equation}

\noindent
This reward function essentially favors frames for inclusion in the summary whose corresponding feature vectors $x_{t^\prime}$ are as close as possible to the centers of the aforementioned clusters.

\subsubsection{Text Saliency Reward ($R_{sal}$)}
\label{textual_reward}

For the text saliency reward, we use the pre-trained AREDSUM model~\cite{AREDSUM}, which takes as input the video transcript and predicts the saliency score (importance) of each sentence $s_i$. 

We define the transcript saliency reward as:

\[R_{sal}(\mathcal{S}) = \sum_{i=1}^M\sum_{j=1}^{T_i}\frac{F_{sal}(s_i)}{T_i}\cdot 1_{\{f_{j}\in\mathcal{S}\}},\]

\noindent
where $M$ is the number of sentences in the transcript, $T_i$ is the number of frames the sentence $s_i$ corresponds to, $F_{sal}(s_{i})$ is the saliency score of sentence $s_i$, and $1_{\{f_{j} \in \mathcal{S}\}}$ is an indicator function that equals $1$ if the $j$-th frame in the sequence of frames corresponding to $s_i$ is included in $\mathcal{S}$, and $0$ otherwise~\footnote{We assume that $F_{sal}(s_i)$ is evenly distributed across the $T_i$ frames in which it corresponds to.}. This reward function encourages the agent to prioritize video segments with brief, information-rich sentences.

The AREDSUM model input is limited to 512 tokens. Thus, in order to handle longer input sequences, we apply a sliding window technique with a window size of $512$ tokens and a sliding step of $256$ tokens. However, since with the aforementioned setup, two windows overlap, the tokens will receive two different saliency values from each window evaluation. To address this issue, we consider the final saliency value of each token to be the average of the saliency values from the two window evaluations.

\subsection{Multimodal Transformer}
\begin{figure}
    \centering
    \includegraphics[width=\linewidth]{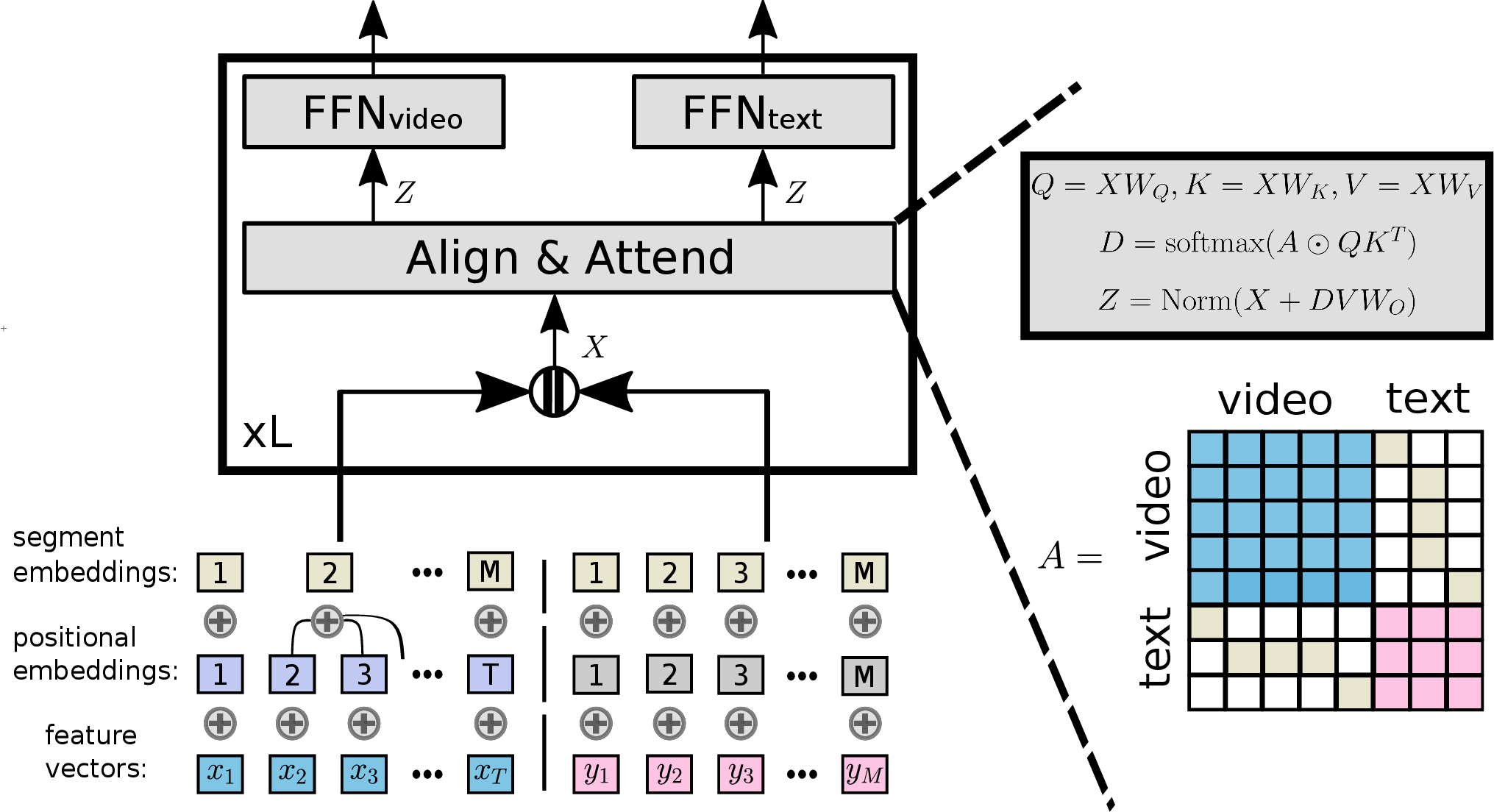}
    \caption{The internal structure of the multimodal Transformer in our solution. See the Align \& Attend module that illustrates how the video and text modalities are aligned.}
    \label{fig:align_and_attend}
\end{figure}
The inputs to the multimodal transformer are the $C$-dimensional feature vectors extracted by the S3D \cite{s3d} and the RoBERTa \cite{roberta} pre-trained models for the frames and the transcript sentences, respectively (see Fig.~\ref{fig:overview}). Learnable positional encodings are then added to these feature vectors to encode positional information within the sequence.
Additionally, learnable segment embeddings are assigned to each sentence and shared with the corresponding frames. These embeddings are then added to the sum of the feature vectors along with the positional encodings.
Finally, the resulting feature vectors from all frames and sentences 
are concatenated together and fed into the \textit{Align \& Attend} module (see Fig. \ref{fig:align_and_attend}).

The \textit{Align \& Attend} module \cite{he2023a2summ} \cite{attention} 
lets features from one modality attend to their corresponding features in the other modality. If the model applied standard self-attention across all features, the differing nature of the modalities could introduce noise due to interactions between unrelated elements. 
To mitigate this, we define an attention mask $\mathbf{A} \in \mathbb{R}^{(T+M) \times (T+M)}$, initialized with all zeros,
where we set the positions that correspond to aligned frames with sentences to \num{1}.

To allow intra-modality attention, features extracted from frames should be able to attend to all other frame features, and features extracted from sentences should attend to all other sentence features. We achieve this by setting the entries at $\mathbf{A}[1:T, 1:T]$ (frame-to-frame attention) and $\mathbf{A}[T:T+M, T:T+M]$ (sentence-to-sentence attention) to \num{1}.
For cross-modal attention between frames and sentences (inter-modality attention), we enable attention only between aligned frame-sentence pairs.
Specifically, if the $i^{th}$ sentence $s_i$ corresponds to frames 
$\{f_i\}_{i\in[k, l]}$, we set 
$\mathbf{A}[T+i, k:l] = 1$ and $\mathbf{A}[k:l,T+i] = 1$.
This ensures that the model captures meaningful cross-modal interactions while minimizing the influence of irrelevant sentences and background frames.

The \textit{Align \& Attend} module is followed by two specialized feedforward layers, 
one for video features ($\text{FFN}_{\text{video}}$) and one for text features ($\text{FFN}_{\text{text}}$) \cite{he2023a2summ}, ~\cite{modality_experts}, which enhance modality-specific processing. 
Finally, the outputs of the two feedforward layers are normalized and added to their respective inputs.
The entire multimodal transformer block is repeated $L$ times, and the $\text{FFN}_{\text{video}}$ layer in the last block outputs the contextualized embeddings $\{h_t\}^T_{t=1}$ that are given as input to the $\text{FFN}_{\text{head}}$ (see~Fig.~\ref{fig:overview}).

\subsection{Training with Policy Gradient}

The goal is to optimize a deep-learning model so that the learned policy $\pi_\theta$, parameterized by $\theta$, selects frames that maximize the expected reward: $J(\theta)=\mathbb{E}_{\pi_\theta}[R(\mathcal{S})]$.

To achieve this, we use the REINFORCE algorithm~\cite{REINFORCE} which updates the model parameters $\theta$ using gradient descent:
\begin{equation} \label{eq:paramupdate}
\theta \leftarrow  \theta - \alpha \nabla_\theta (- J + \beta_1 L_\text{\%} + \beta_2 L_\text{2}),
\end{equation}
where $L_{\%} = ||\frac{1}{T} \sum_{t=1}^T p_t - \epsilon||^{2}$ is a regularization term 
ensuring that approximately $\epsilon \times T$ frames are selected, with $\epsilon$ controlling the target fraction of selected frames, and $L_{2} = \sum \theta^{2}$ is an L2 regularization term 
that prevents overfitting. Furthermore, $\alpha$ is the learning rate, and $\beta_1$, and $\beta_2$ are hyperparameters to balance the influence of $L_{\%}$ and $L_{2}$, respectively~\cite{vsummReinforce}. Since the expected reward $J(\theta)$ is computed over multiple episodes, we approximate its gradient ($\nabla_\theta J$) using Monte Carlo sampling over $K$ episodes:

\begin{equation} \label{eq:reinforce_minus_baseline}
\nabla_\theta J (\theta) \approx \frac{1}{K} \sum_{k=1}^K \sum_{t=1}^T (R_k - b) \nabla_\theta \log \pi_\theta (a_t | h_t),
\end{equation}
where $R_k$ is the reward obtained in the $k^{th}$ episode, $b$ is a moving average of past rewards, which helps reduce variance in gradient estimates, and $a_t$ is the action (frame selection decision) taken at time step $t$, given the frame representation $h_t$, and the $\log \pi_\theta (a_t | h_t)$ term comes from the policy gradient theorem in~\cite{REINFORCE}, which updates $\theta$ to increase the probability of selecting high-reward frames.

\section{Experiments and Results}

In this section, we evaluate the performance of the proposed model for video summarization and highlight detection, both with and without leveraging transcript information. Additionally, we compare our model to DR-DSN~\cite{vsummReinforce} (our baseline), an RL-based approach that uses a biLSTM as its core architecture and relies exclusively on visual information from the video.

\subsection{Setup}
The Transformer model has a context window of \num{50000} frames and \num{50000} sentences. The frame feature size is 1,024, while the sentence feature size is 768. The hidden size $C$ is set to 128, and training is conducted over 60 epochs using K=5 episodes per iteration. The Adam optimizer \cite{adam} is employed with a learning rate of $1 \times 10^{-5}$. Moreover, dropout rates were used and configured as follows: \num{0.1} for the attention layer, $\text{FFN}_{\text{video}}$ and $\text{FFN}_{\text{text}}$, and \num{0.5} for the $\text{FFN}_{\text{head}}$. The AREDSUM model was pre-trained on the CNN/DailyMail dataset \cite{CNNNYT}. Videos were sampled at \num{1} frame per second, ensuring one visual feature per second. Finally, following~\cite{vsummReinforce}, we set the hyperparameters $\lambda$ and $\beta_2$ to \num{20} and $1\times 10^{-5}$, respectively, while we tuned the $\beta_1$ to $0.12$. All training was performed on an Nvidia Tesla V100 GPU with 26 GB VRAM.

\subsection{Datasets}
The model is trained using an unsupervised approach, allowing the training dataset to consist of raw, unannotated videos, with ground-truth annotations required only for evaluation. For this purpose, we use the HowTo100M dataset \cite{miech19howto100m} for training and the MR.HiSum dataset \cite{mrhisum} for evaluation, including their transcripts. The original HowTo100M dataset contains 136 million instructional videos from YouTube. However, due to hardware and time constraints, we randomly selected a subset of 34,000 videos.

The evaluation dataset consists of \num{32}k videos, along with user engagement statistics taken from YouTube. These statistics reflect user engagement throughout each video and can serve as the ground-truth highlight scores of the video frames. Due to the unavailability of many videos, we used a subset of \num{12}k videos for evaluation.
\subsection{Evaluation Metrics}
\subsubsection{Video Summarization Evaluation}
For the task of video summarization, following previous work \cite{he2023a2summ, clover, susinet, UMT, clipit}, \cite{vsummReinforce}, we utilized the F1 score to evaluate how well the generated summaries match the ground-truth summaries. 

However, as highlighted by \cite{rethinkingEvaluation}, the reliability of the F1 score is heavily influenced by the segmentation step, to the extent that even a random method can achieve better results than the state-of-the-art models. To address this, \cite{rethinkingEvaluation} proposed using rank statistics (Spearman's $\rho$ \cite{spearman} and Kendall $\tau$ \cite{kendall}) as alternative metrics. These metrics compare the frame-level rankings obtained from the predicted and ground-truth scores, avoiding the segmentation step.
\subsubsection{Highlight Detection Evaluation}
For the task of highlight detection, following \cite{mrhisum}, we adopt the Mean Average Precision at $\rho\%$ (MAP$\rho$) metric, which evaluates the model's ability to locate the most salient parts of the video. This metric measures how effectively the model ranks the top-$\rho\%$ of frames (as determined by the ground-truth scores) in high positions in the predicted frame-level ranking.
\subsection{Results and Discussion}
Table~\ref{tab:results} presents the performance of the models across various metrics for video summarization and highlight detection. For the MAP metric, we selected values for $\rho=\{5\%, 15\%, 50\%\}$.

We observe that the multimodal Transformer outperforms the unimodal model w.r.t. Spearman’s $\rho$, Kendall’s $\tau$\footnote{The values of Kendall $\tau$ and Spearman's $\rho$ may appear small, but their magnitude is aligned with \cite{rethinkingEvaluation}.}, MAP50, and MAP15, indicating that incorporating textual information enhances performance. However, the unimodal model performs better in terms of the F1 score and MAP5. This suggests that while textual features provide useful context for summarization and highlight detection, visual features remain crucial in certain scenarios.

To understand why the unimodal model outperforms the multimodal model in terms of the MAP5 score, we need to examine this metric further. In essence, MAP5 evaluates the model’s ability to rank the top 5\% of frames at the highest positions. These frames represent a very small portion of the video, often corresponding to a single, localized event. In such cases, the visual modality is highly effective in capturing these moments, while the textual modality may introduce noise. Conversely, for MAP50, where the top 50\% of frames are distributed throughout the video and span multiple events, the textual modality plays a more significant role in influencing viewer engagement. 

Regarding the F1 score (see also earlier remarks and \cite{rethinkingEvaluation}), we applied the KTS algorithm \cite{KTS}, which segments the video based solely on visual content. Consequently, this segmentation approach inherently favors the unimodal model, making direct comparisons with the multimodal model potentially unfair.

Moreover, we observe that the DR-DSN model \cite{vsummReinforce} underperforms in all tasks. This result demonstrates that the integration of the Transformer architecture and the use of textual modality significantly improve model performance. 

Finally, it is evident that the proposed multimodal model with 5 layers ($L$=5) performs better at highlight detection, while the 1-layer variant is more effective for video summarization. However, the opposite is true for the unimodal model: the 5-layer variant excels in video summarization, whereas the 1-layer version performs better at highlight detection. \footnote{Note that an exhaustive exploration of different values for $L$ was explored, but only those that achieved the highest scores for each task are documented.}

\begin{table*}
    \centering
    \caption{Performance comparison of different models on the Mr. HiSum Dataset. The first three columns represent video summarization results, while the last three columns correspond to highlight detection results.}
    \renewcommand{\arraystretch}{1.3}
    \setlength{\tabcolsep}{10pt}
    \begin{tabular}{|l|ccc|ccc|}
        \hline
        \multicolumn{1}{|c|}{} & \multicolumn{3}{c|}{Video Summarization} & \multicolumn{3}{c|}{Highlight Detection} \\
        \cline{2-7}
        & \textbf{F1} & \textbf{Spearman’s $\rho$} \footnotemark[3]  & \textbf{Kendall $\tau$} \footnotemark[3] & \textbf{MAP50} & \textbf{MAP15} & \textbf{MAP5} \\
        \hline
        \textbf{DR-DSN \cite{vsummReinforce}} & 53.28 & 0.0329 & 0.0225 & 56.52 & 23.65 & 20.70 \\
        \hline
        \textbf{Proposed-Unimodal}$_{L=5}$ & \textbf{55.80*} & 0.0348 & 0.0286 & 55. 97 & 28.58 & 25.19 \\
        \textbf{Proposed-Unimodal}$_{L=1}$ & 53.78 & 0.0319 & 0.0214 & 57.38 & 28.95 & \textbf{25.27} \\
        \hline
        \textbf{Proposed-Multimodal}$_{L=5}$ & 54.50 & 0.0636 & 0.0433 & \textbf{59.26} & \textbf{31.37} & 20.70 \\
        \textbf{Proposed-Multimodal}$_{L=1}$ & 54.31 & \textbf{0.0752} & \textbf{0.0514} & 58.71 & 29.70 & 19.32 \\
        \hline
    \end{tabular}
    \label{tab:results}
\end{table*}

\section{Conclusion and Future Work}
This study explores the potential of unsupervised multimodal models for video summarization and highlight detection. We developed a pipeline that, leveraging a reinforcement learning framework, detects highlights and generates summaries that are diverse and representative and include video segments whose corresponding transcript parts carry significant meaning. Our findings indicate that the incorporation of video transcripts enhances the model performance in video summarization, as measured by rank correlation coefficient metrics. Additionally, this approach improves the model’s ability to localize the lower-tier highlights distributed across the video. However, it is less effective at identifying the most salient top-tier highlights, which are often brief and localized events. 

For future work, experiments can be expanded to fully utilize the HowTo100M dataset. Given that the dataset comprises videos from various domains, creating separate domain-specific subsets could also allow for training distinct models tailored to each domain, thereby optimizing performance. Moreover, the model could be extended to process frame-level captions and the audio stream of the video. Lastly, additional datasets, such as~\cite{koupaee2018wikihow} and~\cite{cognimuse}, can be used to further evaluate model performance.

\balance

\bibliographystyle{IEEEtran}

\bibliography{references}

\end{document}